\documentclass[a4paper,fleqn]{cas-dc}
\usepackage[authoryear]{natbib}

\usepackage{placeins}
\usepackage{xcolor}
\usepackage{hyperref}
\usepackage{tabularx}
\usepackage{booktabs}
\usepackage{enumitem}
\usepackage[table]{xcolor}
\hypersetup{
  colorlinks = true,
  linkcolor  = [rgb]{0.6, 0.0, 0.0}, 
  filecolor  = [rgb]{0.6, 0.0, 0.0}, 
  urlcolor   = [rgb]{0.6, 0.0, 0.0},  
  citecolor  = [rgb]{0.6, 0.0, 0.0},  
}
\definecolor{lightblue}{RGB}{230, 240, 255}

\def\tsc#1{\csdef{#1}{\textsc{\lowercase{#1}}\xspace}}
\tsc{WGM}
\tsc{QE}
\tsc{EP}
\tsc{PMS}
\tsc{BEC}
\tsc{DE}

\begin{document}
\let\WriteBookmarks\relax
\def\floatpagepagefraction{1}
\def\textpagefraction{.001}
\let\printorcid\relax 
\shorttitle{}
\shortauthors{}

\title [mode = title]{EAGLE: Elevating Geometric Reasoning through
LLM-empowered Visual Instruction Tuning}                      

\author[1]{Zhihao Li}
\ead{zli3446@uwo.ca}
\credit{Writing – original draft, Writing – review \& editing, Methodology, Software}
\cormark[1]

\author[2]{Yao Du}
\ead{duyao10@meituan.com}
\credit{Formal Analysis, Writing – review \& editing}
\cormark[2]

\author[2]{Yang Liu}
\ead{liuyang509@meituan.com}
\credit{Conceptualization}

\author[3]{Yan Zhang}
\ead{yzhang1995@mail.tsinghua.edu.cn}
\credit{Writing – review \& editing}

\author[4]{Yufang Liu}
\ead{yfliu.antnlp@gmail.com}
\credit{Conceptualization}

\author[2]{Mengdi Zhang}
\ead{zhangmengdi02@meituan.com}
\credit{Formal analysis}

\author[2]{Xunliang Cai}
\ead{caixunliang@meituan.com}
\credit{Formal analysis}

\author[1]{Charles Ling}
\ead{charles.ling@uwo.ca}
\credit{Supervision}

\author[1]{Boyu Wang}
\ead{bwang@csd.uwo.ca}
\credit{Writing – review \& editing, Supervision}
\cormark[2]

\affiliation[1]{organization={Department of Computer Science, Western University},
                city={London},
                postcode={N6A 5B7}, 
                state={Ontario},
                country={Canada}}
\affiliation[2]{organization={Meituan Inc.},
                city={Beijing},
                postcode={100102}, 
                country={China}}
\affiliation[3]{organization={Department of Automation, Tsinghua University},
                city={Beijing},
                postcode={100084}, 
                country={China}}
\affiliation[4]{organization={School of Computer Science and Technology, East China Normal University},
                city={Shanghai},
                postcode={200062}, 
                country={China}}

\cortext[cor1]{Work done during an internship at Meituan.}
\cortext[cor2]{Corresponding authors.}

\begin{abstract}
Multi-modal Large Language Models (MLLMs) have advanced greatly in general tasks. However, they still face challenges in geometric reasoning, a task that requires synergistic integration of visual recognition proficiency and complex reasoning strength. Existing MLLMs prioritize optimizing the LLM backbone to enhance problem-solving capabilities, while rarely emphasizing improvements in discerning visual elements. However, we observe that MLLMs suffer from severe visual perception deficiencies in geometric domains, including inaccurate geometric comprehension and severe visual hallucinations, which constrain their reasoning performance. To address this issue, we revisit geometric reasoning through a visual-centric lens that highlights the role of visual perception. To achieve this, we propose \textbf{EAGLE}, a coarse-to-fine visual adaptation framework for geometric reasoning that progressively adapts the visual pathway to geometric perception. Specifically, given the substantial disparity between geometric diagrams and natural images, we first introduce \textbf{Geometric Knowledge Injection}. This process explores fundamental knowledge from diagram-caption data to enhance recognition capabilities and improve geometry-language alignments. Then, recognizing that different elements contribute unequally in the reasoning process, we introduce \textbf{Geometric Knowledge Refinement}. This stage leverages LLM-driven chain-of-thought solutions to provide additional reasoning supervision for visual refinement, fostering a synergistic interplay between visual comprehension and geometric reasoning. Finally, we develop EAGLE, a geometry expert with improved perception and reasoning capabilities. Extensive experiments demonstrate its effectiveness on four popular benchmarks. Code will be available at \url{https://github.com/zhli-cs/EAGLE}.
\end{abstract}


\maketitle
\section{Introduction}\label{intro}
Large Language Models (LLMs) have demonstrated remarkable performance across a variety of tasks \citep{zou2025novel,liao2025attack}. Latest LLMs even exhibit human-level reasoning capabilities in complicated tasks such as mathematical problem-solving \citep{imani2023mathprompter, zhu2024distilling} and code generation \citep{sun2024unicoder,chai2025xcot}. These studies mainly focus on textual contexts, while multi-modal mathematical problems \citep{zheng2024multimodal,gllava} that require extra visual comprehension proficiency remain relatively underexplored. Due to the intrinsic limitation of LLMs in \emph{visual perception}, they are not well-suited for solving such problems. Consequently, Multi-modal Large Language Models (MLLMs), which extend LLMs by incorporating an additional vision encoder, have attracted substantial attention. 

\begin{figure}
\centering
\includegraphics[width=\linewidth]{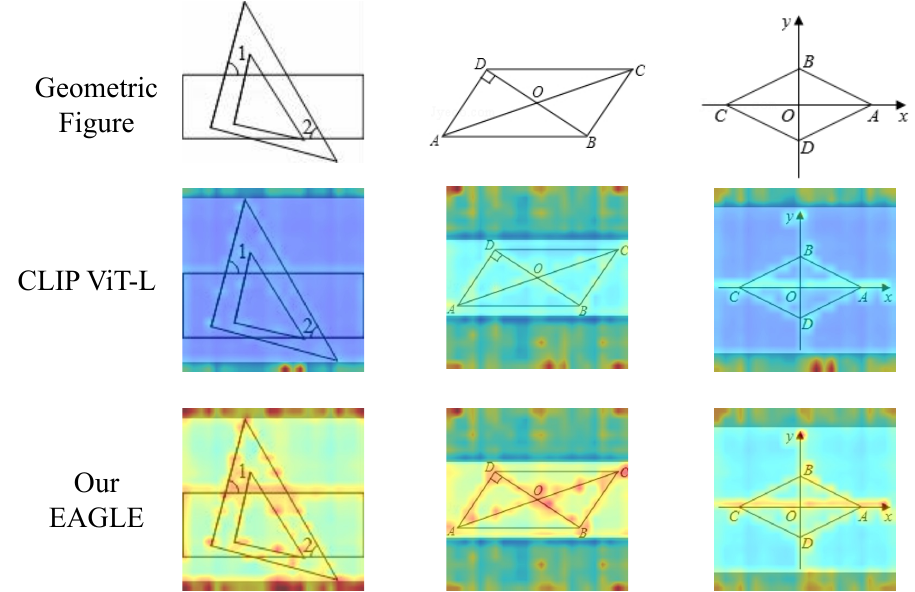}
\caption{Comparison of attention maps between CLIP ViT and EAGLE. Regions highlighted with warm colors denote areas with higher model attention, whereas regions with cold colors indicate lower attention. It can be seen that CLIP ViT struggles to capture geometric shapes.}
\label{attn_visualization}
\vspace{-5ex}
\end{figure}

MLLMs have demonstrated strong performance on general multi-modal tasks \citep{xu2015show,antol2015vqa}. However, they still face challenges in complex reasoning. In particular, Geometry Problem Solving (GPS), a challenging task that requires the synergy of precise visual perception and rigorous mathematical deduction, has emerged as a valuable testbed. Some works \citep{math-llava,gllava} have made preliminary efforts by fine-tuning LLMs on geometric data. However, these studies mainly focus on improving the reasoning capacity of the LLM backbone, overlooking the reliability of the visual comprehension, which is typically supported by the standard CLIP ViT \citep{CLIP}. Nevertheless, due to the substantial domain gap between CLIP's natural image pre-training and geometric diagrams, we find that CLIP's inadequate geometric recognition capability is another primary bottleneck, and our empirical analysis substantiates this deficiency. Specifically, as visualized in Figure \ref{attn_visualization} and Figure \ref{caption_cases}, CLIP exhibits insufficient perception of geometric elements and fails to accurately identify fundamental elements (e.g., points, lines, angles). This perceptual ambiguity propagates to the language model, leading to severe visual geometric hallucinations and inaccurate descriptions. Such inadequacies limit MLLMs' problem-solving capability, underscoring the urgency to adapt vision encoders to the geometric domain.

Conventional visual optimization strategies often adapt the vision encoder separately from the target MLLM. For example, MAVIS~\citep{MAVIS} fine-tunes the CLIP ViT on geometric image-caption pairs using the CLIP text encoder as a proxy supervision space. While such decoupled optimization can improve fundamental geometric perception, the absence of the target LLM during visual adaptation may limit the alignment between visual representations and the downstream LLM. In geometric problem solving, different visual components contribute unequally to the solution process, making it desirable for visual representations to be refined according to reasoning-relevant information. Some recent works address this issue through joint visual-language optimization. For instance, MathGLM-Vision~\citep{mathglm} jointly adapts the visual pathway during large-scale mathematical SFT, demonstrating the practical value of visual adaptation for multimodal mathematical reasoning. However, how visual adaptation should be delicately structured across different learning objectives for geometric reasoning remains underexplored.

To address the aforementioned issue, we disentangle the geometric adaptation into a coarse-to-fine visual enhancement framework, and organize it into two stages: geometric perception acquisition and subsequent reasoning-guided refinement. We propose \textbf{EAGLE}, which \textbf{E}lev\textbf{A}tes \textbf{G}eometric reasoning through \textbf{L}LM-\textbf{E}mpowered visual instruction tuning. The core idea is to progressively adapt the vision encoder to the linguistic space defined by the LLM and subsequently refine it with CoT-guided supervision. Our training pipeline consists of two stages: \textbf{Geometric Knowledge Injection} (Geo-KI) and \textbf{Geometric Knowledge Refinement} (Geo-KR). In the first stage, we enhance visual understanding of basic geometric elements using diagram-caption data. This process not only equips the vision encoder with fundamental geometric knowledge but also establishes adaptive geometry-language alignments with the LLM backbone. Moreover, recognizing that different elements play inherently unequal roles in deriving solutions, we further refine the vision encoder in the second stage using additional reasoning supervision. This subtle optimization is guided by LLM's step-by-step problem-solving solutions. Specifically, we leverage question-answer data with chain-of-thought (CoT) rationales. During this refinement, the vision encoder undergoes further adaptation guided by CoT supervision. Note that generic visual optimization like LLaVA-NeXT \citep{li2024llavanext-ablations} and ShareGPT4V \citep{sharegpt4v} straightforwardly unlocks the vision encoder without considering domain gaps. However, our ablation studies demonstrate that such simplistic strategies yield only marginal gains or even induce performance degradation. Finally, we present EAGLE, a 7B-scale specialist that exhibits improved geometric perception and competitive reasoning performance. Our contributions are as follows: 
\begin{figure}
\centering
\includegraphics[width=\linewidth]{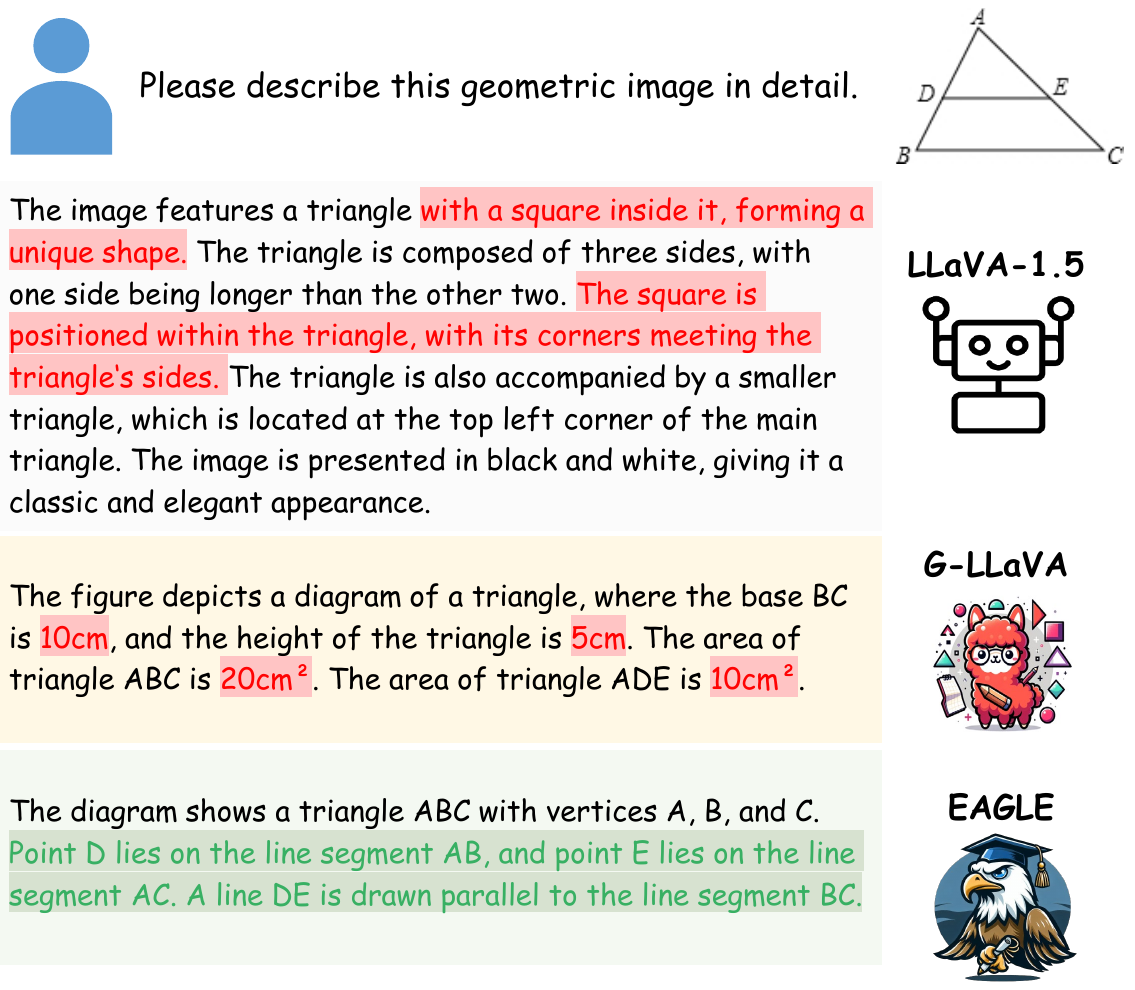}
\caption{Comparison of the geometric caption performance. EAGLE provides more accurate geometric descriptions (green background) while LLaVA-1.5 and G-LLaVA face severe visual hallucinations (red background).}
\label{caption_cases}
\vspace{-5ex}
\end{figure}
\begin{itemize}
    \item We provide a geometry-specific empirical analysis of visual perception deficiencies in existing MLLMs, showing how inaccurate recognition of geometric elements and hallucinated visual content can constrain downstream reasoning. 
    \item We introduce a visual adaptation framework for geometric reasoning that progressively enhances perception proficiency. Specifically, Geometric Knowledge Injection (Geo-KI) enables ViT to acquire fundamental geometric knowledge and facilitates geometry-language alignment. Subsequently, Geometric Knowledge Refinement (Geo-KR) leverages LLM-driven Chain-of-Thought rationale to provide additional reasoning supervision for visual refinement.
    \item We develop the geometric specialist model EAGLE, which exhibits competitive performance on popular benchmarks with only 7B parameters. We also conduct extensive quantitative and qualitative experiments to explore the impact of different visual adaptation strategies.
\end{itemize}

The remainder of this paper is organized as follows. Section~\ref{related work section} reviews related work. Section~\ref{method section} elaborates on the proposed visual enhancement framework. Section~\ref{experiment section} presents extensive experiments and in-depth analyses. Finally, Section~\ref{conclusion section} concludes the paper.

\section{Related Work}\label{related work section}
\subsection{Multi-modal Large Language Models}
Recent years have witnessed thriving developments for large language models \citep{bai2022training,touvron2023llama,touvron2023llama2}, which display remarkable performance across various natural language processing tasks. Concurrently, many researchers attempt to harness the prowess of LLMs in diverse multi-modal scenarios, leading to the prosperity of multi-modal large language models \citep{GPT-4V(vision),llava,mplug-owl2,nie2025vision,zeng2025dual}. Modern MLLMs typically comprise a vision encoder that extracts visual clues (e.g., CLIP \citep{CLIP}), a cross-modal projector that aligns visual and textual features, and a language foundation model that serves as the cognitive core (e.g., LLaMA \citep{touvron2023llama}). In early studies, MLLMs primarily focus on conventional vision-language tasks, as demonstrated by  BLIP-2 \citep{li2023blip} and Mini-GPT4 \citep{minigpt4}. Recently, in light of the remarkable advancements achieved by LLMs in complex reasoning \citep{yu2024thought,yuan2024advancing}, there has been a surge of research focusing on multi-modal reasoning \citep{wang2024exploring,wang2024stop}. For example, Zheng \textit{et al.} \citep{zheng2023ddcot} devised a novel DDCoT prompting technique that decouples LLMs' reasoning and recognition capabilities, enabling the transfer of these advancements to multi-modal contexts. He \textit{et al.} \citep{he2024multi} proposed to improve the multi-modal CoT capability through latent space learning. Despite these significant achievements, we observe that MLLMs still face challenges in effectively addressing multi-modal mathematical problems involving geometric elements.

\subsection{Geometry Problem Solving}
Geometry problem solving \citep{jia2024describe,Math-puma,gllava,xia2025geox} is a challenging reasoning task that requires MLLMs to possess proficiency in both visual perception and mathematical problem-solving. Given the rapid development of complex reasoning, many methods have achieved marked developments. For example, G-LLaVA \citep{gllava} constructs a geometric dataset Geo170K and develops a geometric expert model. Math-LLaVA \citep{math-llava} develops a math-focused MLLM trained on a curated MathV360K dataset. GeoGen~\citep{GeoGen} improves reasoning supervision by synthesizing symbolically grounded CoT data and further
integrating symbolic verification during inference. These studies mainly focus on enhancing the LLMs' reasoning strength while neglecting the deficiency of visual recognition. Several methods have been proposed to mitigate this issue. For instance, MAVIS \citep{MAVIS} developed a math-specific MAVIS-Caption dataset to fine-tune ViT through contrastive learning. MultiMath~\citep{multimath} adopts visual and mathematical instruction tuning, followed by process-supervised reinforcement learning. MathGLM Vision~\citep{mathglm} performs large-scale mathematical SFT on the diverse MathVL corpus while activating the visual encoder, providing complementary evidence for the importance of visual adaptation in mathematical reasoning. However, these studies do not explicitly disentangle the visual adaptation process. In contrast, we propose a progressive coarse-to-fine framework that decomposes visual enhancement into geometric perception acquisition and CoT-guided refinement, together with systematic analyses of different visual adaptation strategies.

\begin{figure*}[t!]
\centering
\includegraphics[width=\linewidth]{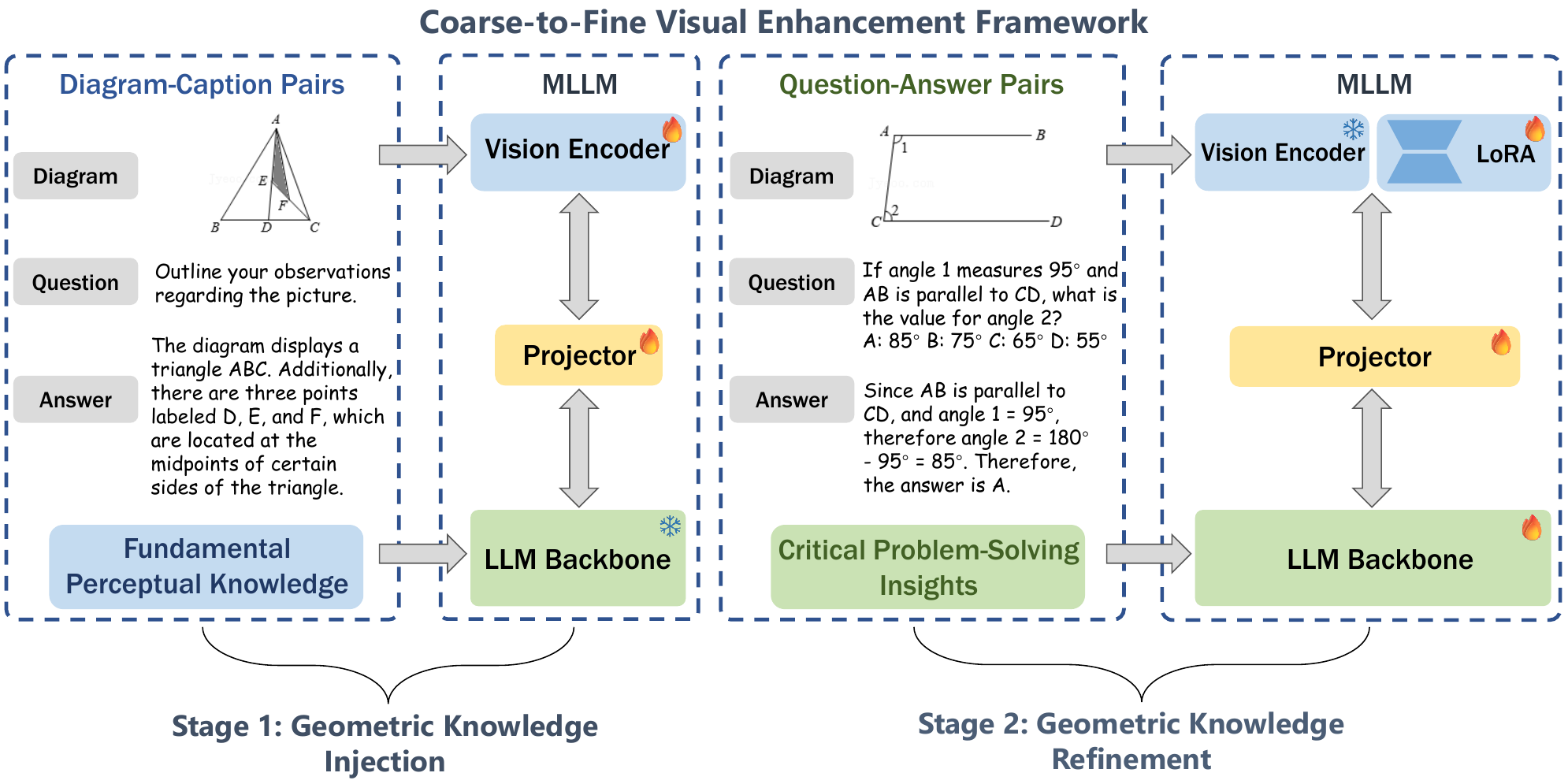}
\caption{The overview of our proposed coarse-to-fine visual enhancement framework. During Geometric Knowledge Injection, fundamental geometric knowledge is extracted from diagram-caption data, enhancing the recognition capabilities of the vision encoder. During Geometric Knowledge Refinement, critical problem-solving insights are derived from LLM-driven step-by-step question-answer pairs, providing additional reasoning supervision for visual refinement.}
\label{method_architecture}
\end{figure*}
\section{Method}\label{method section}
In this section, we describe EAGLE in detail, as shown in Figure \ref{method_architecture}. Our methodological contribution lies in a two-stage visual adaptation strategy designed for geometric reasoning that progresses from geometric perception enhancement to further visual refinement. Given that reasoning is more challenging than general tasks~\citep{zhang2025learning,ke2025a}, we decouple geometric adaptation into a two-stage coarse-to-fine process. The term ``coarse-to-fine'' characterizes the progressive granularity of visual knowledge and supervision across the two stages. Specifically, the first stage is Geometric Knowledge Injection (Geo-KI), where the vision encoder acquires geometry-specific primitives and relations that are largely absent from natural-image pre-training. The second stage is Geometric Knowledge Refinement (Geo-KR), where the acquired visual representations are further refined according to their relevance to specific reasoning processes. Both stages optimize the visual pathway within the MLLM, with the Geo-KI Stage adapting it to the fixed linguistic space and the Geo-KR Stage further refining it with CoT-guided supervision.

\subsection{Stage I: Geometric Knowledge Injection}
The CLIP ViT used in MLLMs exhibits severe deficiencies due to the domain gap between geometric and natural images. To mitigate this issue, we focus on injecting fundamental geometric knowledge (e.g., elements, shapes, and correlations) into the vision encoder. The goal is to adapt the vision encoder to geometric domains and establish groundwork for more advanced reasoning. Specifically, we utilize 60K diagram-caption pairs to fine-tune ViT in an end-to-end manner, with the CLIP ViT perceiving images and the frozen LLM providing a fixed linguistic space for visual adaptation. We also optimize the cross-modal projector between the vision encoder and the LLM to foster adaptive geometry-language alignments. It is noteworthy that our Geometric Knowledge Injection differs from existing fine-tuning strategies that decouple ViT from MLLMs and optimize it separately \citep{MAVIS}, which may introduce extra discrepancies when aligning with LLMs. Differently, we integrate visual optimization into an end-to-end MLLM training pipeline. This process fine-tunes ViT and fosters geometric vision-language alignment simultaneously, which equips the model with superior geometric expertise. To summarize, we optimize the vision encoder and the cross-modal projector during this stage. The LLM backbone is kept frozen to provide a fixed linguistic space during visual adaptation.

\subsection{Stage II: Geometric Knowledge Refinement}
After the knowledge injection stage, the vision encoder is equipped with fundamental geometric knowledge and is capable of describing geometric diagrams reliably. However, geometric reasoning poses greater challenges than captioning tasks, as it not only demands precise recognition but also requires the effective use of visual information during reasoning. To enhance this capability, we leverage the LLM-driven step-by-step problem-solving process to further refine the vision encoder. This process is also achieved in an end-to-end MLLM training pipeline, where 110K question-answer pairs with detailed solution steps are employed for visual instruction tuning. Specifically, we incorporate LoRA \citep{lora} modules into the vision encoder to enable visual refinement under CoT supervision, while we keep the main body of the ViT frozen to maintain its prior geometric perceptual knowledge. Moreover, we unfreeze the LLM backbone to improve its reasoning strength, which continually provides reasoning-guided supervision for visual refinement. In this stage, as the LLM absorbs knowledge from CoT rationales, the LoRA-based vision encoder receives additional learning signals from intermediate reasoning tokens, promoting in-depth geometric comprehension. We note that standard instruction tuning \citep{math-llava,gllava} typically focuses on improving LLMs with a frozen ViT, while our Knowledge Refinement surpasses them by leveraging CoT solutions to provide additional reasoning supervision to the visual pathway, leading to superior geometric comprehension. To summarize, we employ LoRA for the vision encoder while keeping the cross-modal projector and the LLM backbone trainable during this stage.

\subsection{Mechanistic Analysis of Geo-KR}\label{Mechanistic Analysis of Geo-KR}
To illustrate how CoT supervision refines the visual pathway, we provide additional mechanistic analysis. Specifically, we explicitly formulate the gradient propagation during the Geometric Knowledge Refinement stage. Let $I$ denote the input geometry diagram, $S_{\mathrm{in}}$ the input instruction/question, and $S_{\mathrm{tar}} =\{S_{\mathrm{tar}}^{t}\}_{t=1}^{l}$ the target response of length $l$. During Geo-KR, $S_{\mathrm{tar}}$ consists of the CoT reasoning sequence followed by the final answer.

Let $V_{\phi}(I)$ denote the visual tokens produced by the vision encoder, where $\phi$ represents its parameters. The visual tokens are mapped into the language embedding space through the cross-modal projector
$P_{\psi}$:
\begin{equation}
Z = P_{\psi}\left(V_{\phi}(I)\right),
\end{equation}
where $\psi$ denotes the parameters of the projector and $Z$ is the projected visual representation.

Given the preceding target tokens $S_{\mathrm{tar}}^{(<t)} =\{S_{\mathrm{tar}}^{1},\ldots,S_{\mathrm{tar}}^{t-1}\}$, the CoT-based training objective is
\begin{equation}
\mathcal{L}_{\mathrm{CoT}} = -\sum_{t=1}^{L}\log p_{\theta}\!\left(S_{\mathrm{tar}}^{t} \,\middle|\, S_{\mathrm{tar}}^{<t},\,S_{\mathrm{in}},\,Z\right),
\end{equation}
where $\theta$ denotes the parameters of the LLM.

Let $\ell_t$ denote the token-level negative log-likelihood at step $t$.
The gradient propagated to the vision encoder can then be expressed as
\begin{equation}
\nabla_{\phi}\mathcal{L}_{\mathrm{CoT}}=
\sum_{t=1}^{l}\frac{\partial \ell_t}{\partial Z}\frac{\partial Z}{\partial V_{\phi}(I)}\frac{\partial V_{\phi}(I)}{\partial \phi}.
\end{equation}

During training, gradients from losses on intermediate reasoning tokens propagate through the LLM and cross-modal projector to the LoRA-adapted vision encoder. Compared with answer-only supervision, CoT provides additional token-level learning signals along the reasoning trajectory. This gradient pathway remains valid even when the LLM parameters are frozen, as gradients can still propagate through the LLM to the visual tokens. These additional reasoning signals can therefore support further refinement of the visual pathway during training.

\subsection{Model Training}
To ensure consistency and reduce potential biases in the geometry-centered visual enhancement framework, we employ end-to-end training at both stages rather than optimizing the vision encoder separately. We utilize an autoregressive language modeling loss as the training objective. Specifically, to generate the $t^{th}$ target token, the geometry diagram $I$ is first encoded by the vision encoder $V_{\phi}$ and mapped through the cross-modal projector $P_{\psi}$ to obtain the projected visual representation $Z=P_{\psi}(V_{\phi}(I))$. The projected visual representation $Z$, the input sentence $S_{\mathrm{in}}$, and the preceding target tokens $S_{\mathrm{tar}}^{(<t)}$ are then used for next-token prediction by the LLM parameterized by $\theta$. The training objective is formulated as follows:
\begin{equation}
\mathcal{L}(S_{\mathrm{tar}},S_{\mathrm{in}},I)
=
-\sum_{t=1}^{l}
\log p_{\theta}
\left(
S_{\mathrm{tar}}^{t}
\mid
S_{\mathrm{tar}}^{(<t)}, S_{\mathrm{in}}, Z
\right),
\end{equation}
where $S_{\text{tar}}$ denotes the target sentence, $l$ denotes the length.
\begin{table}[h]
\centering
\caption{Hyper-parameter settings. ``Geo-KI'' and ``Geo-KR'' indicate Geometric Knowledge Injection and Geometric Knowledge Refinement, respectively.}
\begin{tabularx}{\columnwidth}{
    @{} 
    >{\hspace{0.5em}}l 
    >{\centering\arraybackslash}X 
    >{\centering\arraybackslash}X 
    @{}
}
\toprule
 &  Geo-KI  & Geo-KR    \\ \hline
Learning Rate (LR)                       & 1e-5                                                            & 3e-5                                                                                \\ 
LR Schedule                        & \multicolumn{2}{c}{cosine decay}                    \\
LR Warmup Ratio         &  \multicolumn{2}{c}{0.03}  \\
Epoch    & 1    &   2    \\
Optimizer      &     \multicolumn{2}{c}{AdamW}     \\
LoRA Alpha & -- & 16 \\
LoRA Rank    &  --    & 64     \\
LoRA Dropout  &  -- & 0.05 \\
\bottomrule
\end{tabularx}
\label{hyper-parameters}
\end{table}

\begin{table}[t!]
\caption{Computational cost of EAGLE. Batch-size denotes the per-GPU batch size, and memory denotes the peak memory usage.}
\centering
\small
\setlength{\tabcolsep}{3pt}
\begin{tabular}{ccccc}
\toprule
Stage & GPUs & Total GPU Hours & Batch-size & GPU Memory \\
\midrule
Geo-KI & 1 & 3.86 & 6 & 30.68 GB \\
Geo-KR & 4 & 15.63 & 6 & 65.65 GB \\
\bottomrule
\end{tabular}
\label{computational_cost}
\end{table}

\begin{table*}[t!]
\caption{Comparison on the GeoQA benchmark. Note that $^\dagger$ indicates reproduced methods.}
\centering
\setlength{\tabcolsep}{0pt}
\begin{tabularx}{\textwidth}{
    @{} 
    >{\hspace{0.5em}}l 
    >{\hsize=1.4\hsize\centering\arraybackslash}X 
    >{\hsize=0.8\hsize\centering\arraybackslash}X 
    >{\hsize=0.8\hsize\centering\arraybackslash}X 
    @{}
}
\toprule
Methods    & LLM Backbone  & LLM Size            & Accuracy \\ \hline
\multicolumn{4}{c}{\textit{Heuristics Baselines}} \\ \hline
Random Chance & -    & -         & 25.0                              \\
Frequent      & -    & -         & 32.1                              \\ \hline
\multicolumn{4}{c}{\textit{Conventional Models}} \\ \hline
Geoformer \citep{UniGeo}  & - & -     & 46.8                              \\
UniMath \citep{Unimath}   & - & -      & 50.0                              \\ \hline
\multicolumn{4}{c}{\textit{General MLLMs}}   \\ \hline
mPLUG-Owl2 \citep{mplug-owl2} &  LLaMA-2 &7B   & 15.7                            \\
ShareGPT4V \citep{sharegpt4v} & Vicuna-1.5 &7B   & 16.2                            \\
LLaVA-1.5 \citep{llava1.5} & Vicuna-1.5 &7B    & 21.1                              \\
LLaVA-1.5 \citep{llava1.5}  & Vicuna-1.5 &13B   & 23.2                             \\
LLaVA-NeXT \citep{li2024llavanext-ablations} & LLaMA-3 &8B   & 33.9                             \\ 
Qwen2-VL \citep{qwen2-vl} & Qwen2 & 7B & 46.2 \\
Qwen2.5-VL \citep{qwen2.5} & Qwen2.5 & 3B & 56.8 \\
InternVL2.5 \citep{chen2024internvl} & QLLaMA & 8B & 58.1 \\
GPT-4o \citep{gpt4o} & - & - & 61.4 \\  \hline
\multicolumn{4}{c}{\textit{Mathematical MLLMs}}  \\       \hline              
Math-LLaVA \citep{math-llava} & Vicuna-1.5 &13B & 47.8 \\
Math-PUMA \citep{Math-puma}  & DeepSeek-Math &7B   & 61.8 \\
Math-PUMA \citep{Math-puma}  & Qwen2 &7B   & 63.6 \\
G-LLaVA \citep{gllava}   & LLaMA-2 &7B    & 64.2                              \\
InternLM-XC2 \citep{dong2024internlm} & InternLM2 &7B      & 66.4                              \\
MAVIS w/o DPO \citep{MAVIS}   & MAmmoTH-2 &7B      & 66.7                              \\
GeoX \citep{xia2025geox}   & Geo-LLM &7B    & 67.4 \\
G-LLaVA$^\dagger$ \citep{gllava}   & LLaMA-3.1 &8B    & 67.6   \\
MAVIS \citep{MAVIS}   & MAmmoTH-2 &7B      & 68.3                              \\ 
G-LLaVA$^\dagger$ \citep{gllava}   & Qwen-2.5 &7B    & 71.5    \\
\rowcolor{lightblue}
\textbf{EAGLE}  & Vicuna-1.5 &7B       & 67.1                              \\ 
\rowcolor{lightblue}
\textbf{EAGLE}  & LLaMA-3.1 &8B       & 70.4                              \\
\rowcolor{lightblue}
\textbf{EAGLE}  & Qwen-2.5 &7B       & \textbf{75.4}                              \\\bottomrule
\end{tabularx}
\label{GeoQA_results}
\end{table*}
\section{Experiment}\label{experiment section}
In this section, we present a comprehensive evaluation of EAGLE through extensive experiments on four benchmarks covering both geometric reasoning and fine-grained geometric perception. Beyond standard performance metrics, we conduct in-depth ablation studies to rigorously verify the individual contribution and efficacy of each proposed training strategy. Finally, we provide detailed qualitative visualizations and specific case studies, which further demonstrate the superior capabilities of the proposed coarse-to-fine visual enhancement framework in geometric perception and complex problem-solving.

\subsection{Experiment Settings}
\noindent\textbf{Datasets.}
We utilize the geometry-specific Geo170K dataset introduced by G-LLaVA \citep{gllava} to train our model. Geo170K is constructed utilizing the training sets of GeoQA+ \citep{GeoQA+} and Geometry3K \citep{Geometry3K}, which consists of two parts: 60K examples of geometric alignment data primarily in the form of image-caption pairs, and 110K examples of geometric question-answer data with detailed CoT rationales. 

\noindent\textbf{Implementation Details.}
We develop EAGLE with the LLaVA-1.5 architecture \citep{llava1.5}, which comprises a CLIP ViT-L/14 \citep{CLIP} as the vision encoder and a 7B Vicuna-1.5 \citep{chiang2023vicuna} as the LLM backbone. We employ a two-layer MLP as the cross-modal projector to connect the vision encoder and the base LLM. The input image resolution is set to 336 by 336 pixels. The detailed training hyper-parameter settings are presented in Table \ref{hyper-parameters}. Specifically, to avoid overfitting, we follow G-LLaVA \citep{gllava} to train EAGLE for one epoch in the geometric knowledge injection stage and two epochs in the geometric knowledge refinement stage. We employ the AdamW optimizer \citep{adamw} during the training process. The learning rate is set to $1e^{-5}$ and $3e^{-5}$ on each stage with a cosine decay learning rate scheduler. Geo-KI is conducted with one NVIDIA H100 80GB GPU, while Geo-KR utilizes four H100 GPUs. The detailed computation cost is reported in Table~\ref{computational_cost}.

\noindent\textbf{Evaluation and Metrics.}
We evaluate our model on four popular benchmarks, namely GeoQA \citep{GeoQA}, MathVista \citep{mathvista}, We-Math \citep{wemath}, and Geoperception~\citep{zhang2024euclid}. GeoQA is a commonly used geometry dataset that mainly focuses on plane geometry. MathVista encompasses various mathematical tasks that require visual comprehension, where we select the Geometry Problem Solving (GPS) task within the MathVista testmini subset for evaluation. We-Math explores the underlying problem-solving principles during the reasoning process beyond performance. Geoperception evaluates MLLMs' ability to accurately transcribe 2D geometric information from an image. All models are assessed in a zero-shot setting. Our model is specifically prompted to generate responses in a fixed format, and the predicted answers are automatically extracted via a regular expression. We note that Geo170K includes the training split of GeoQA+, while GeoQA uses a separate test split. Although they share similar data distributions, we follow the standard evaluation protocol of prior geometry MLLMs and additionally evaluate on MathVista, We-Math, and Geoperception for broader assessment.

\subsection{Main Results.}
\noindent\textbf{Comparisons on GeoQA.} 
As shown in Table \ref{GeoQA_results}, EAGLE with Vicuna-1.5-7B backbone achieves 67.1\% accuracy on GeoQA, improving over the closely related G-LLaVA baseline (64.2\%) and slightly outperforming contemporary mathematical-centric approaches, such as Math-PUMA~\citep{Math-puma} (trained on 996K mathematical data) and MAVIS w/o DPO \citep{MAVIS} (trained on more than 1.4M mathematical examples), despite being trained on a substantially smaller dataset comprising only 170K examples. We further instantiate EAGLE with stronger LLMs to examine its scalability. With LLaMA-3.1-8B and Qwen-2.5-7B, EAGLE achieves 70.4\% and 75.4\%, respectively, achieving competitive performance among recent geometric MLLMs. These results show that EAGLE maintains competitive performance under a relatively lightweight geometric training setting and further benefits from stronger language backbones. Importantly, we note that since these models differ in their foundation LLMs and training resources, we do not interpret the absolute performance differences as direct evidence of methodological superiority. Instead, the matched-backbone comparisons in Table~\ref{controlled_backbone} more directly isolate the contribution of the proposed visual adaptation strategy.

\begin{table*}[t!]
\caption{Comparison on testmini set of the MathVista benchmark for Geometry Problem Solving. Note that $^\dagger$ indicates reproduced methods.}
\centering
\setlength{\tabcolsep}{0pt}
\begin{tabularx}{\textwidth}{
    @{} 
    >{\hspace{0.5em}}l 
    >{\hsize=1.4\hsize\centering\arraybackslash}X 
    >{\hsize=0.8\hsize\centering\arraybackslash}X 
    >{\hsize=0.8\hsize\centering\arraybackslash}X 
    @{}
}
\toprule
Methods      & LLM Backbone  &  LLM Size          & Accuracy \\ \hline
\multicolumn{4}{c}{\textit{Heuristics Baselines}} \\ \hline
Random Chance  & -   & -          & 21.6                             \\
Frequent       & -  & -           & 34.1                              \\
Human     & -    & -           & 48.4                              \\ \hline
\multicolumn{4}{c}{\textit{General MLLMs}}                \\ \hline
LLaVA-1.5 \citep{llava1.5}& Vicuna-1.5 &7B   & 17.3                             \\
LLaVA-1.5 \citep{llava1.5}  & Vicuna-1.5 &13B   & 18.8                             \\
InstructBLIP \citep{dai2023instructblip} & Vicuna &7B    & 20.7                   \\
ShareGPT4V \citep{sharegpt4v} & Vicuna-1.5 &7B   & 22.1                             \\
SPHINX-V1 \citep{sphinx}   & LLaMA-2 &13B       & 23.1                              \\
mPLUG-Owl \citep{mplug-owl} &  LLaMA &7B       & 23.6                              \\
Gemini 1.0 Nano2 \citep{gemini}  & -  & -           & 23.6                              \\
LLaVA-NeXT \citep{li2024llavanext-ablations}  & LLaMA-3 &8B   & 25.0  \\
LLaVAR \citep{llavar}   & Vicuna &13B     & 25.0                              \\
MiniGPT4 \citep{minigpt4} & LLaMA-2 &7B        & 26.0                              \\
Qwen-VL-Plus \citep{qwen-vl-plus} &  Qwen &7B         & 38.5                              \\
Gemini 1.0 pro \citep{gemini} & -  & -        & 40.4                              \\ 
Qwen2.5-VL \citep{qwen2.5} & Qwen2.5 & 3B & 42.2 \\
Qwen2-VL \citep{qwen2-vl} & Qwen2 & 7B & 49.0 \\
GPT-4V \citep{GPT-4V(vision)} & -  & -        & 50.5                              \\\hline
\multicolumn{4}{c}{\textit{Mathematical MLLMs}}         \\ \hline
Deepseek-VL \citep{lu2024deepseek}  & DeepSeek-LLM &7B        & 28.4                              \\
VCAR \citep{jia2024describe}   & Vicuna-1.5 &7B   & 34.6 \\
Math-PUMA \citep{Math-puma}  & DeepSeek-Math &7B   & 39.9 \\
Qwen2-VL \citep{qwen2-vl}   &  Qwen2 &7B   & 40.9 \\
Math-PUMA \citep{Math-puma}   & Qwen2 &7B   & 48.1 \\
G-LLaVA \citep{gllava}    & LLaMA-2 &7B   & 53.4                              \\ 
G-LLaVA$^\dagger$ \citep{gllava}   & LLaMA-3.1 &8B    & 57.7 \\
MAVIS w/o DPO \citep{MAVIS}   & MAmmoTH-2 &7B      & 63.2                              \\
MAVIS \citep{MAVIS}   & MAmmoTH-2 &7B      & 64.1                              \\
G-LLaVA$^\dagger$ \citep{gllava}   & Qwen-2.5 &7B    & 64.9 \\
\rowcolor{lightblue}
\textbf{EAGLE}    & Vicuna-1.5 &7B       & 54.3                              \\ 
\rowcolor{lightblue}
\textbf{EAGLE}  & LLaMA-3.1 &8B       & 60.1                              \\
\rowcolor{lightblue}
\textbf{EAGLE}  & Qwen-2.5 &7B       & \textbf{66.8}                              \\\bottomrule
\end{tabularx}
\label{MathVista_results}
\end{table*}

\noindent\textbf{Comparisons on MathVista.}
As shown in Table \ref{MathVista_results}, EAGLE with Vicuna-1.5-7B achieves 54.3\% accuracy in Geometry Problem Solving and outperforms GPT-4V by 3.8\% on the testmini split of MathVista. It also exhibits competitive performance compared with other baseline models. For example, it exceeds its baseline model LLaVA-1.5-7B with a clear margin of 37\% and the seminal G-LLaVA-7B model by 0.9\%. Notably, EAGLE also surpasses larger 13B models like SPHINX-V1 \citep{sphinx} and LLaVAR \citep{llavar}. Similar to GeoQA, EAGLE benefits from stronger language backbones. The LLaMA-3.1-8B and Qwen-2.5-7B variants achieve 60.1\% and 66.8\%, respectively. These results indicate that the proposed framework can effectively scale with stronger language backbones. Similarly, since these variants employ different foundation models, we rely on the controlled comparisons in Table~\ref{controlled_backbone}, rather than the above results alone, to assess the contribution of the proposed visual adaptation strategy.

\begin{table*}[t!]
\caption{Comparison on the Geoperception benchmark. POL: PointLiesOnLine, POC: PointLiesOnCircle, ALC: AngleClassification, LHC: LineComparison, PEP: Perpendicular, PRA: Parallel, and EQL: Equals.}
\vspace{1em}
\centering
\small
\setlength{\tabcolsep}{8pt}
\begin{tabular}{lccccccc}
\toprule
Methods & POL$\uparrow$ & POC$\uparrow$ & ALC$\uparrow$ & LHC$\uparrow$ & PEP$\uparrow$ & PRA$\uparrow$ & EQL$\uparrow$ \\
\midrule
G-LLaVA & 2.42 & 4.77 & 55.78 & 32.47 & 0.24 & 1.42 & 0.97 \\
\rowcolor{lightblue}
\textbf{EAGLE} & \textbf{13.14} & \textbf{5.48} & \textbf{58.14} & \textbf{35.08} & \textbf{1.29} & \textbf{5.26} & \textbf{1.93} \\
\bottomrule
\end{tabular}
\label{geoperception_results}
\end{table*}

\noindent\textbf{Comparisons on Geoperception.} We further evaluate fine-grained geometric perception on Geoperception. As shown in Table~\ref{geoperception_results}, EAGLE consistently outperforms G-LLaVA across all seven perception tasks. The improvements are particularly pronounced for recognizing point-line relations and parallel structures, while consistent gains are also observed for angle classification, line comparison, perpendicularity, and equality recognition. These results provide direct quantitative evidence that the proposed visual adaptation effectively enhances the recognition of fundamental geometric elements and relations.

\begin{table*}[t]
\caption{Comparison on the We-Math testmini set with four-dimensional metrics (IK: Insufficient Knowledge, IG: Inadequate Generalization, CM: Complete Mastery, RM: Rote Memorization) of two criteria (Loose and Strict). Avg. denotes the average score. The best and second-best scores in each category are marked in \textbf{bold} and \underline{underlined}, respectively.}
\centering
\setlength{\tabcolsep}{0pt}
\begin{tabularx}{\textwidth}{@{} 
    >{\hsize=1.5\hsize\raggedright\arraybackslash}X  
    >{\hsize=1.5\hsize\centering\arraybackslash}X    
    *{4}{>{\hsize=0.7\hsize\centering\arraybackslash}X} 
    >{\hsize=1.5\hsize\centering\arraybackslash}X   
    *{4}{>{\hsize=0.7\hsize\centering\arraybackslash}X} 
@{}}
\toprule
        Methods & Avg. (Loose)$\uparrow$ & IK$\downarrow$ & IG$\downarrow$ & CM$\uparrow$ & RM$\downarrow$ & Avg. (Strict)$\uparrow$ & IK$\downarrow$ & IG$\downarrow$ & CM$\uparrow$ & RM$\downarrow$ \\
        \midrule
        DeepSeek-VL-7B  & 20.95 & 69.14 & \textbf{4.57} & 18.67 & \textbf{28.99} & \underline{6.29} & 69.14 & \textbf{4.57} & \underline{4.00} &\textbf{84.78}  \\
        G-LLaVA-13B  & \underline{22.29} & \textbf{64.19} & \textbf{4.57} & \underline{20.00} & 35.98 & \textbf{6.48} & \textbf{64.19} & \textbf{4.57} & \textbf{4.19} & 86.59  \\
        \rowcolor{lightblue}
        \textbf{EAGLE-7B} & \textbf{22.86} & \underline{66.10} & \textbf{4.57} & \textbf{20.57} & \underline{29.87} & \textbf{6.48} & \underline{66.10} & \textbf{4.57} & \textbf{4.19} & \underline{85.71} \\
        \bottomrule
\end{tabularx}
\label{wemath}
\end{table*}

\noindent\textbf{Comparisons on We-Math.} To investigate the underlying reasoning principles of mathematical reasoning, we employ the We-Math benchmark with its granular four-dimensional metrics, as presented in Table~\ref{wemath}. We note that the reported G-LLaVA results are taken from the official We-Math leaderboard and correspond to the 13B model, whereas EAGLE uses a 7B backbone. We observe that EAGLE outperforms DeepSeek-VL-7B~\citep{lu2024deepseek} across most metrics. Moreover, despite its smaller model size, EAGLE exhibits superior performance compared with G-LLaVA-13B \citep{gllava} in the average loose score while remaining competitive under the rigorous strict criterion. Collectively, these results demonstrate that our coarse-to-fine enhancement strategy boosts reasoning reliability across diverse evaluation standards.

\begin{table}[t!]
\caption{Controlled comparison under matched LLM backbones. Note that $^\dagger$ denotes our reproduction of G-LLaVA using the same LLM backbone and geometric training data as EAGLE. Results are reported as mean $\pm$ standard deviation over five runs with different random seeds.}
\vspace{1em}
\centering
\small
\setlength{\tabcolsep}{2pt}
\begin{tabular}{lcccc}
\toprule
Method & Backbone & Data & GeoQA & MathVista-GPS \\
\midrule

G-LLaVA$^\dagger$ & Vicuna-1.5-7B  & Geo170K & 64.4$\pm$0.4 & 52.9$\pm$0.2 \\
\rowcolor{lightblue}
\textbf{EAGLE}     & Vicuna-1.5-7B  & Geo170K & \textbf{67.1$\pm$0.2} & \textbf{54.3$\pm$0.3} \\
\midrule

G-LLaVA$^\dagger$ & LLaMA-3.1-8B & Geo170K & 67.6$\pm$0.3 & 57.7$\pm$0.5  \\
\rowcolor{lightblue}
\textbf{EAGLE}     & LLaMA-3.1-8B  & Geo170K & \textbf{70.4$\pm$0.4} & \textbf{60.1$\pm$0.2} \\
\midrule

G-LLaVA$^\dagger$ & Qwen2.5-7B & Geo170K & 71.5$\pm$0.2 & 64.9$\pm$0.3 \\
\rowcolor{lightblue}
\textbf{EAGLE}     & Qwen2.5-7B & Geo170K & \textbf{75.4$\pm$0.2} & \textbf{66.8$\pm$0.4} \\
\bottomrule
\end{tabular}
\label{controlled_backbone}
\end{table}

\noindent\textbf{Controlled comparison with G-LLaVA.} To disentangle the effect of visual adaptation from the choice of the LLM backbone, we conduct controlled comparisons with G-LLaVA$^\dagger$ under three language backbones, namely Vicuna-1.5, LLaMA-3.1~\citep{grattafiori2024llama} and Qwen2.5~\citep{qwen2.5}, as shown in Table~\ref{controlled_backbone}. For each backbone, both methods use the same geometric training data and training configuration, while differing primarily in the visual adaptation strategy: G-LLaVA$^\dagger$ keeps the vision encoder frozen throughout training, whereas EAGLE employs FFT for geometric perception acquisition and LoRA for CoT-guided refinement. The results averaged over five runs show that EAGLE consistently outperforms the G-LLaVA baseline across the three LLM backbones, with relatively small variability across runs. These results further demonstrate the generalizability of the proposed visual adaptation strategy across language backbones.

\subsection{Comparison with Recent Mathematical and Geometric MLLMs}
We further contextualize EAGLE with several recent geometric MLLMs, including MultiMath~\citep{multimath}, MathGLM-Vision~\citep{mathglm}, and GeoGen-SFT~\citep{GeoGen}, as summarized in Table~\ref{recent_comparison}. Since these methods differ substantially in their foundation models, geometric fine-tuning data, and training paradigms, we report their results primarily to provide a broader reference to recent progress rather than as strict comparisons.

Specifically, MultiMath leverages a math-specialized backbone together with multi-stage instruction tuning and reinforcement learning, while MathGLM-Vision performs large-scale mathematical SFT with visual encoder adaptation. GeoGen-SFT further expands Geo170K with symbolically generated geometric data. To summarize, these baselines all incorporate several geometric datasets during optimization, while in contrast, EAGLE relies only on Geo170K for geometric fine-tuning and focuses specifically on the design of visual adaptation. It is also noteworthy that EAGLE consistently benefits from a stronger language backbone. Replacing Vicuna-1.5-7B with Qwen2.5-7B improves the performance from 67.1\% to 75.4\% on GeoQA and from 54.3\% to 66.8\% on MathVista-GPS. Despite using only Geo170K for geometric fine-tuning, the Qwen2.5-based EAGLE achieves performance comparable to MultiMath-7B on MathVista-GPS and approaches GeoGen-SFT-7B on GeoQA. These results indicate that the proposed coarse-to-fine visual adaptation can effectively benefit from more powerful LLMs while maintaining a relatively lightweight geometric training setting. 

\begin{table*}[t!]
\caption{Comparison with recent strong mathematical and geometric MLLMs.
Results of competing methods are quoted from the corresponding papers and are provided as reference due to differences in foundation models and geometric fine-tuning data. EAGLE further benefits from a stronger LLM, with the Qwen2.5-7B variant achieving competitive performance using only Geo170K.}
\centering
\small
\setlength{\tabcolsep}{1.5pt}
\renewcommand{\arraystretch}{1.05}

\begin{tabularx}{\textwidth}{@{}
>{\raggedright\arraybackslash\hsize=0.75\hsize}X
>{\centering\arraybackslash\hsize=0.95\hsize}X
>{\centering\arraybackslash\hsize=1.75\hsize}X
>{\centering\arraybackslash\hsize=0.60\hsize}X
>{\centering\arraybackslash\hsize=0.92\hsize}X
@{}}
\toprule
Methods & Backbone & Geometric Fine-tuning Data & GeoQA & MathVista-GPS \\ 
\midrule
MultiMath      & DeepSeekMath-RL-7B & Geo170K, MultiMath300K, MathV360K & --   & 66.8  \\
MathGLM-Vision & GLM-4V-9B          & MathVL (Geo170K, GeomVerse, UniGeo-Calculation, etc.) & --   & 64.4 \\
GeoGen-SFT     & Qwen2.5-VL-7B      & Geo170K, GeoExpand, GeoSynth & \textbf{78.0} & \textbf{74.0}  \\ 
\midrule
\rowcolor{lightblue}
\textbf{EAGLE} & Vicuna-1.5-7B      & Geo170K & 67.1 & 54.3  \\ 
\rowcolor{lightblue}
\textbf{EAGLE} & Qwen2.5-7B          & Geo170K & 75.4 & 66.8  \\
\bottomrule
\end{tabularx}
\label{recent_comparison}
\end{table*}
\begin{table}[t!]
\caption{Further analysis with different visual enhancement strategies. ``--'' indicates that the corresponding stage is omitted entirely, whereas ``Freeze'' indicates that the stage is performed with the vision encoder frozen.} ``FFT'' denotes Full Fine-tuning. Results are obtained on the GeoQA benchmark.
\centering
\setlength{\tabcolsep}{0pt}
\begin{tabularx}{\columnwidth}{@{} >{\centering\arraybackslash}X >{\centering\arraybackslash}X >{\centering\arraybackslash}X >{\centering\arraybackslash}X @{}}
\toprule
Methods &  Geo-KI   & Geo-KR &    Accuracy  \\ \midrule
G-LLaVA       &  Freeze   & Freeze & 64.2 \\
1                        & --                                                          & LoRA                                                       &  63.8                           \\ 
2                        & --                                                            & FFT                                                       & 64.3                           \\ 
3                        & --                                                  & Freeze                                             & 64.6                           \\
\rowcolor{lightblue}
\textbf{EAGLE}                     & FFT                                                 & LoRA                                                       & \textbf{67.1}                           \\ \bottomrule
\end{tabularx}
\label{Ablations}
\end{table}
\subsection{Further Analysis}
\noindent\textbf{Analysis of Geometric Knowledge Injection.}
The Geo-KI stage leverages diagram-caption pairs to equip the MLLM with fundamental geometric knowledge, laying the groundwork for subsequent visual refinement. To rigorously validate its necessity, we systematically exclude it from the training pipeline and assess its impact on final performance, as shown in Table \ref{Ablations}. We observe that Methods 1–3 consistently underperform EAGLE, highlighting the critical role of knowledge injection in acquiring basic geometric expertise. Notably, directly applying LoRA refinement without Geo-KI (Method 1) yields only 63.8\% accuracy, indicating that LoRA is more effective for refining previously acquired geometric representations than for directly adapting a generic vision encoder from scratch.
\begin{table}[t!]
\caption{Generalization of EAGLE under different input resolutions.}
\centering
\small
\setlength{\tabcolsep}{8pt}
\begin{tabular}{ccc}
\toprule
Methods &Input Resolution & GeoQA \\
\midrule
\textbf{EAGLE} &336\;$\times$\;336 & 67.1 \\
\textbf{EAGLE}&448\;$\times$\;448 & \textbf{67.7} \\
\bottomrule
\end{tabular}
\label{resolution_comparison}
\end{table}

\noindent\textbf{Analysis of Geometric Knowledge Refinement.}
We also investigate the impact of CoT-guided Knowledge Refinement, as presented in Table \ref{Ablations}. Notably, Methods 1 to 3 demonstrate performance comparable to or exceeding that of the baseline G-LLaVA 7B solely via Geometric Knowledge Refinement, highlighting the effectiveness of Geo-KR supervision for downstream geometric reasoning. Notably, Methods 1 and 2 underperform Method 3 despite updating the ViT with LoRA or full fine-tuning during Geo-KR. We argue that without the Geo-KI stage, ViT has not yet acquired geometry-specific perceptual knowledge, and directly refining it with QA supervision may introduce less reliable visual signals for subsequent reasoning. Consequently, freezing the ViT in Method 3 yields slightly better performance. We note that Method 3 also performs comparably to G-LLaVA even without the Geo-KI stage. The reason stems from the fact that G-LLaVA freezes the ViT and mainly updates the cross-modal projector and LLM backbone, while its insufficient geometric perception capability limits its performance. Together, these results provide further evidence that adapting the ViT with diagram-caption alignment data is essential before applying reasoning-guided refinement.

\noindent\textbf{Generalization to Higher Resolution.}
Geometric diagrams often contain thin lines, small symbols, and fine-grained textual annotations, making input resolution particularly relevant. We therefore further evaluate EAGLE under a higher-resolution $448\times448$ setting. As shown in Table~\ref{resolution_comparison}, EAGLE achieves 67.7\% on GeoQA at $448\times448$, indicating that the proposed framework remains effective when extended to a higher-resolution visual setting.

\noindent\textbf{Impact of Visual Adaptation on Natural-Image Performance.} We further evaluate the vision encoder on ImageNet-1K to examine whether the two-stage adaptation affects CLIP's original visual capability. As shown in Table~\ref{imagenet_results}, Geo-KI leads to some degradation on natural-image classification, which can arise when fully fine-tuning a pretrained vision encoder toward a specialized geometric domain, as the representation is shifted from generic natural-image features toward geometry-specific visual patterns. In contrast, the subsequent LoRA-based Geo-KR does not cause any additional performance drop and instead slightly recovers the performance. These results suggest that the adapted encoder retains most of its natural-image classification capability despite some performance degradation following geometric adaptation.

\begin{table}[t!]
\caption{ImageNet-1K classification performance of the vision encoder before and after the two-stage visual adaptation.}
\centering
\small
\setlength{\tabcolsep}{6pt}
\begin{tabular}{lcc}
\toprule
Model & Top-1 Accuracy & Top-5 Accuracy \\
\midrule
Original CLIP & \textbf{74.29} & \textbf{93.74} \\
\rowcolor{lightblue}
\textbf{EAGLE} after Geo-KI & 63.81 & 87.67 \\
\rowcolor{lightblue}
\textbf{EAGLE} after Geo-KR & 64.82 & 88.16 \\
\bottomrule
\end{tabular}
\label{imagenet_results}
\end{table}

\begin{table}[t]
\caption{Impact of different visual training strategies. During the Geo-KI stage, the projector remains trainable, while the LLM backbone is kept frozen. In the Geo-KR stage, both the projector and the LLM are set to be trainable. Results are obtained on the GeoQA benchmark. Results are reported as mean $\pm$ standard deviation over five runs with different random seeds.}
\centering
\setlength\tabcolsep{1mm}
\setlength{\tabcolsep}{0pt}
\begin{tabularx}{\columnwidth}{@{} >{\centering\arraybackslash}X >{\centering\arraybackslash}X >{\centering\arraybackslash}X >{\centering\arraybackslash}X @{}}
\toprule
Methods &  Geo-KI   & Geo-KR &    Accuracy  \\ \midrule
1                        & Freeze                                                          & LoRA                                                       &  62.1$\pm$0.4                            \\ 
2                        & LoRA                                                            & LoRA                                                       & 65.1$\pm$0.2                           \\ 
3                        & FFT                                                  & FFT                                             & 64.3$\pm$0.3                           \\
4                        & FFT                                                  & Freeze                                                     & 66.0$\pm$0.3                           \\ 
\rowcolor{lightblue}
\textbf{EAGLE}                     & FFT                                                  & LoRA                                                       & \textbf{67.1$\pm$0.2}                          \\ \bottomrule
\end{tabularx}
\label{Ablation_results}
\end{table}
\begin{table}[t!]
\caption{Comparison between decoupled and end-to-end visual adaptation. Both settings use the same Geo170K data and identical Geo-KR training.}
\vspace{1em}
\centering
\small
\setlength{\tabcolsep}{5pt}
\begin{tabular}{lcc}
\toprule
Stage-1 Visual Adaptation & Geo-KR & GeoQA \\
\midrule
Decoupled CLIP Adaptation & LoRA & 65.3 \\
\textbf{EAGLE} (End-to-End) & LoRA & \textbf{67.1} \\
\bottomrule
\end{tabular}
\label{decoupled_comparison}
\end{table}

\subsection{Ablation Study}
\noindent\textbf{Different Strategies for Knowledge Injection.}
As shown in Table \ref{Ablation_results}, we conduct comparative experiments to investigate different visual enhancement strategies for Knowledge Injection. Comparing EAGLE with Methods 1 and 2, we observe that fine-tuning the entire vision encoder in this stage leads to superior results. We explain that although the CLIP ViT has been trained on an extensive corpus of natural images, its limited exposure to geometry diagrams hinders its recognition and comprehension performance. Consequently, fully fine-tuning the vision encoder significantly facilitates the geometry-centered visual augmentation and establishes the groundwork for a deeper reasoning process. 

\noindent\textbf{Different Strategies for Knowledge Refinement.}
Geo-KR aims to guide visual optimization with step-by-step problem-solving expertise. For deeper comprehension, we evaluate various refining strategies, including fine-tuning the entire vision encoder, freezing the vision encoder, and incorporating additional LoRA modules. As shown in Table \ref{Ablation_results}, EAGLE achieves superior performance by integrating LoRA modules, while fully fine-tuning (Method 3) leads to unsatisfactory results, even poorer than freezing the visual backbone (Method 4). These results support the benefit of LoRA-based visual refinement during Geo-KR.

\noindent\textbf{Comparison with Decoupled Visual Adaptation.} To examine the potential alignment issue introduced by decoupled visual optimization, we construct a counterpart that first adapts the CLIP vision encoder using the CLIP text encoder as a proxy supervision space and then realigns it to the target MLLM through projector fine-tuning. Both settings use the same Geo170K alignment data and identical Geo-KR training. As shown in Table~\ref{decoupled_comparison}, EAGLE achieves 67.1\% on GeoQA, compared with 65.3\% for the decoupled counterpart. This result suggests that directly adapting the visual pathway within the target MLLM can reduce the potential mismatch introduced by intermediate proxy-space optimization and provide a more suitable initialization.

\begin{table}[t!]
\caption{Impact of LLM optimization and CoT supervision during the
Geo-KR stage. Results are obtained on the GeoQA benchmark, where results for EAGLE are reported as mean $\pm$ standard deviation over five runs with different random seeds.}
\centering
\small
\setlength{\tabcolsep}{1.5pt}
\renewcommand{\arraystretch}{1.05}

\begin{tabular}{
    @{}
    >{\raggedright\arraybackslash}m{0.25\columnwidth}
    >{\centering\arraybackslash}m{0.16\columnwidth}
    >{\centering\arraybackslash}m{0.13\columnwidth}
    >{\centering\arraybackslash}m{0.22\columnwidth}
    >{\centering\arraybackslash}m{0.14\columnwidth}
    @{}
}
\toprule
Methods
    & \shortstack[c]{Vision\\Encoder}
    & LLM
    & Supervision
    & Accuracy \\
\midrule

LLaVA-1.5-7B
    & -- & -- & -- & 21.1 \\

LLaVA-1.5-13B
    & -- & -- & -- & 23.2 \\

\textbf{EAGLE$^{\dagger}$}
    & LoRA & Freeze & Answer-only & 29.6$\pm$0.3 \\

\textbf{EAGLE$^*$}
    & LoRA & Freeze & CoT & 31.3$\pm$0.2 \\

\rowcolor{lightblue}
\textbf{EAGLE}
    & LoRA & FFT & CoT & \textbf{67.1$\pm$0.2} \\

\bottomrule
\end{tabular}
\label{freeze LLM}
\end{table}
\noindent\textbf{Performance with a frozen LLM backbone.}
We further demonstrate the efficacy of coarse-to-fine visual enhancement by applying it while freezing the core LLM. Specifically, as shown in Table \ref{freeze LLM}, we freeze the LLM backbone throughout the training pipeline, updating only the vision encoder and projector. To isolate the effect of CoT supervision on visual refinement, we additionally compare answer-only supervision with full CoT supervision while keeping the visual adaptation setting unchanged. We observe that CoT improves GeoQA accuracy from 29.6\% (EAGLE$^{\dagger}$) to 31.3\% (EAGLE$^*$). Together with the gradient analysis in Section~\ref{Mechanistic Analysis of Geo-KR}, this result supports the benefit of additional reasoning supervision for visual refinement, even when the LLM backbone remains frozen.

Moreover, even though EAGLE$^*$ relies entirely on the original reasoning capabilities of the frozen LLM and conducts reasoning without new geometric-specific problem-solving expertise, it still achieves a 10.2 percentage points improvement over the baseline LLaVA-1.5-7B. This result provides further evidence that adapting the visual pathway can improve geometric problem-solving performance with a frozen LLM.

\begin{figure*}[t!]
    \centering
    \includegraphics[width=\linewidth]{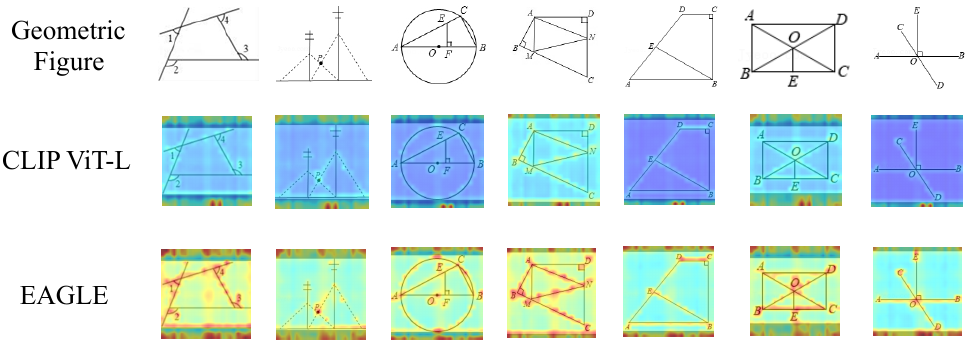}
    \caption{More attention map visualizations between CLIP ViT and EAGLE.}
    \label{visualization}
\end{figure*}
\begin{figure*}
\centering
\includegraphics[width=\linewidth]{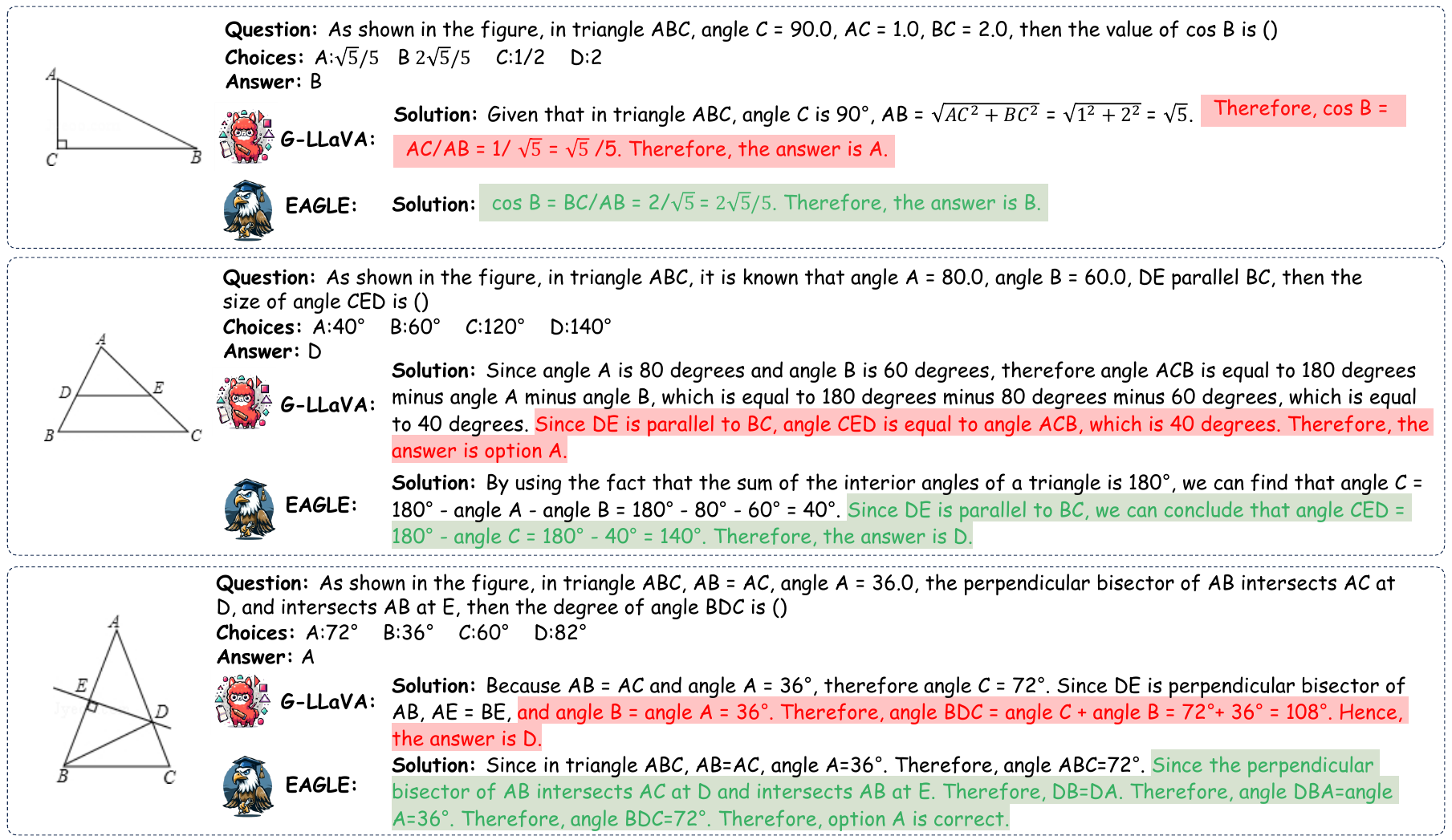}
\caption{Comparison of the geometric problem-solving capability between EAGLE and G-LLaVA on the GeoQA test set. Red background indicates incorrect reasoning, whereas green background highlights correct problem-solving processes.}
\label{qualitative_cases}
\end{figure*}

\subsection{Qualitative Analysis}
\noindent\textbf{More Attention Map Visualizations.}
We provide additional qualitative evidence through attention map visualizations of the vision encoder, further underscoring the enhanced perceptual capability of EAGLE over baseline LLMs that utilize a frozen CLIP ViT. As illustrated in Figure~\ref{visualization}, the original CLIP ViT struggles to discern fine-grained fundamental geometric structures, such as vertices, angles, and outlines. In contrast, empowered by the proposed coarse-to-fine visual enhancement framework, EAGLE effectively mitigates this limitation. These qualitative observations complement the quantitative perception results.

\noindent\textbf{Case Studies on GeoQA.} As illustrated in Figure \ref{qualitative_cases}, we conduct qualitative comparisons between EAGLE and the geometric specialist model, G-LLaVA, on the GeoQA test set. Our analysis reveals that G-LLaVA has severe geometric hallucinations, which frequently fabricates spurious geometric attributes (e.g., non-existent height and angle values) and constructs flawed problem-solving processes that lead to incorrect conclusions. Specifically, in the first case, G-LLaVA exhibits a fundamental disconnect between visual perception and mathematical formulation by incorrectly computing the cosine of angle B as $\cos B=AC/AB$, failing to effectively integrate symbolic visual perception into mathematical reasoning. In the second case, G-LLaVA generates inaccurate diagram descriptions (as previously noted in Figure \ref{caption_cases}), which further compromises its reasoning process and results in incorrect answers. In contrast, EAGLE demonstrates effective synergies between vision perception and geometric reasoning. In these examples, EAGLE produces more accurate geometric descriptions and correct step-by-step solutions.

\begin{figure*}[t]
\centering
\includegraphics[width=\linewidth]{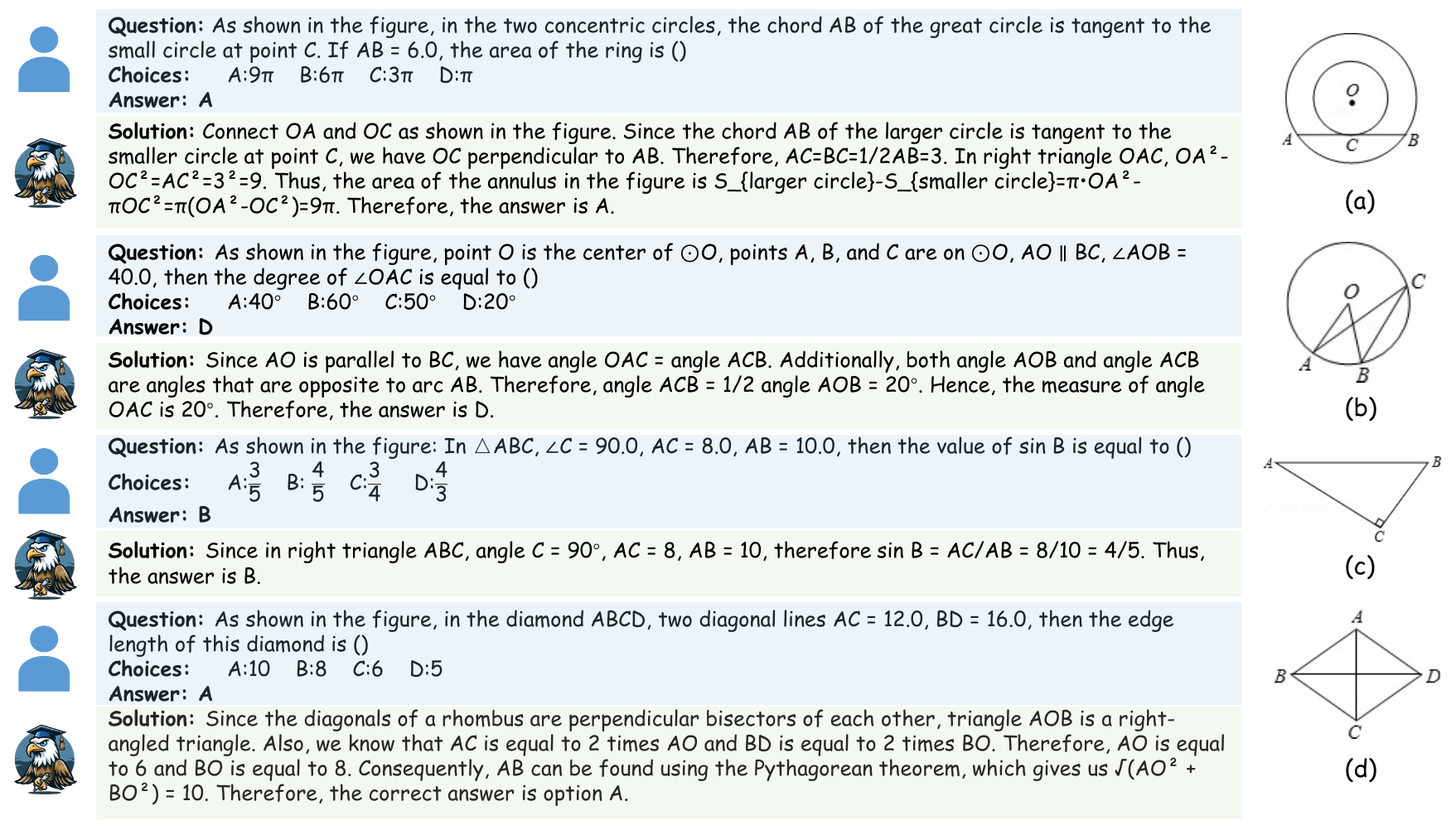}
\caption{Geometric reasoning cases of EAGLE on the MathVista benchmark.}
\label{qualitative_cases_mathvista}
\end{figure*}

\noindent\textbf{Case Studies on MathVista.} We also provide reasoning cases from the testmini set of MathVista regarding Geometry Problem Solving. As shown in Figure \ref{qualitative_cases_mathvista}, EAGLE effectively addresses geometric problems of various types, including calculations of area, angle, length, and trigonometric functions. Notably, in Figure \ref{qualitative_cases_mathvista} (a), EAGLE even demonstrates the proficiency to draw auxiliary lines to assist reasoning, such as connecting OA and OC to form the triangle OAC. These resolution processes further highlight the superior geometric recognition and problem-solving strength of our proposed model. 

\section{Conclusion}\label{conclusion section}
In this paper, we focus on mitigating the visual deficiency of MLLMs to enhance their geometry reasoning capabilities. We provide a geometry-specific empirical analysis showing that inadequate geometric recognition remains an important bottleneck for geometric reasoning. Then, to address this, we introduce a coarse-to-fine visual adaptation framework for geometric reasoning and develop a geometric specialist model, EAGLE. Our approach is structured into two progressive stages: Geometric Knowledge Injection and Geometric Knowledge Refinement. In the first stage, EAGLE acquires fundamental geometric knowledge (e.g., points, lines, and angles) from diagram-caption data, helping it improve geometric comprehension and geometry-language alignment. In the second stage, EAGLE leverages LLM-driven step-by-step CoT solutions to provide reasoning-guided supervision for visual refinement. Through this training process, the visual pathway is progressively adapted for geometric understanding and problem solving. Comprehensive evaluations on popular geometric benchmarks, along with in-depth ablation studies, demonstrate the effectiveness of EAGLE. Beyond geometry, the proposed coarse-to-fine visual adaptation strategy may also extend to other domains with specialized visual distributions, such as tables and data visualizations, which we leave for future investigation.

\printcredits

\section*{Declaration of competing interest}
The authors declare the following personal relationships, which may be considered as potential competing interests: Boyu Wang, given his role as Action Editor, had no involvement in the peer review of this article and had no access to information regarding its peer review. Full responsibility for the editorial process for this article was delegated to another journal editor. If there are other authors, they declare that they have no known competing financial interests or personal relationships that could have appeared to influence the work reported in this paper.

\section*{Acknowledgements}
Z. Li and B. Wang are supported by the Natural Sciences and Engineering Research Council of Canada (NSERC), Discovery Grants program.

\section*{Data availability}
The datasets used in this work are publicly available. The code will be publicly released at \url{https://github.com/zhli-cs/EAGLE}.

\bibliographystyle{cas-model2-names}

\bibliography{reference}

\end{document}